\documentclass{IEEEtran4PSCC}
\usepackage{amsmath,amsfonts}
\usepackage{algorithmic}
\usepackage{algorithm}
\usepackage{textcomp}
\usepackage{stfloats}
\usepackage{url}
\usepackage{verbatim}
\usepackage{graphicx}
\usepackage{cite}
\usepackage{balance}
\usepackage{xcolor}
\usepackage{multirow}
\usepackage{booktabs}
\usepackage{threeparttable}
\usepackage{arydshln}
\renewcommand{\baselinestretch}{0.96}
\usepackage{graphicx}
\usepackage{amsmath}
\makeatletter
\let\old@ps@headings\ps@headings
\let\old@ps@IEEEtitlepagestyle\ps@IEEEtitlepagestyle
\def\psccfooter#1{%
    \def\ps@headings{%
        \old@ps@headings%
        \def\@oddfoot{\strut\hfill#1\hfill\strut}%
        \def\@evenfoot{\strut\hfill#1\hfill\strut}%
    }%
    \def\ps@IEEEtitlepagestyle{%
        \old@ps@IEEEtitlepagestyle%
        \def\@oddfoot{\strut\hfill#1\hfill\strut}%
        \def\@evenfoot{\strut\hfill#1\hfill\strut}%
    }%
    \ps@headings%
}
\makeatother

\psccfooter{%
        \parbox{\textwidth}{\hrulefill \\ \small{24th Power Systems Computation Conference} \hfill \begin{minipage}{0.2\textwidth}\centering \vspace*{4pt} \includegraphics[scale=0.06]{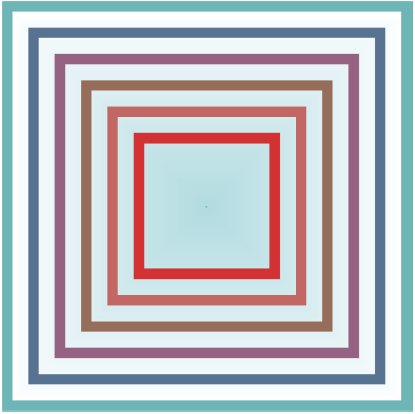}\\\small{PSCC 2026} \end{minipage} \hfill \small{Limassol, Cyprus --- June 8-12, 2026}}%
}

\begin{document}
%

\title{Supervised Reinforcement Learning for the Coordination of Distributed Energy Resources}






\author{
    \IEEEauthorblockN{Haoyuan Deng$^{1,2,\dagger}$, Yihong Zhou$^{3,\dagger}$, Thomas Morstyn$^{3}$, and Yi Wang$^{1,2,*}$}
    \IEEEauthorblockA{$^1$The University of Hong Kong, Hong Kong SAR, China\\
    $^2$The University of Hong Kong Shenzhen Institute of Research and Innovation, Shenzhen, China\\
    $^3$University of Oxford, Oxford, U.K. \\    
    $^{\dagger}$Equal contributions, 
    $^*$Corresponding author
    \vspace{-2mm}
    }
} 


\maketitle

\begin{abstract}
The increasing integration of distributed energy resources (DERs) is crucial for power system decarbonization, yet unlocking DERs' flexibility is challenged by their inherent uncertainties and modelling complexity. As traditional optimization methods struggle with such uncertainty and complexity of DERs, reinforcement learning (RL) has emerged as a promising alternative for DER management. However, standard RL methods suffer from sample inefficiency and sub-optimality when trained from scratch. Inspired by the training paradigms in large language models, this paper proposes a Supervised Reinforcement Learning (SRL) framework for learning DER coordination policies. This framework first pre-trains a policy on demonstration data in a supervised-learning fashion, which is then further fine-tuned using RL. Furthermore, we propose a two-step fine-tuning process: offline fine-tuning for enhancing policy performance and online fine-tuning for adapting it to the real-world dynamics. Experiments demonstrate that RL implementations based on the proposed framework significantly outperform all benchmarks, achieving high cost efficiency even under low-quality demonstration data.
\end{abstract}


\begin{IEEEkeywords}
Accurate Modelling, Distributed Energy Resources (DERs), Reinforcement Learning (RL), Supervised Pre-training, Two-step Fine-tuning.
\end{IEEEkeywords}

\thanksto{The work was supported in part by the National Natural Science Foundation of China (U25B6016), in part by the Research Grants Council of the Hong Kong SAR (HKU 17201225), in part by Advanced Research + Invention Agency (ARIA) under Project SAGEflex: Safeguarded AI Agents for Grid-Edge Flexibility, and in part by Engineering and Physical Sciences Research Council (EPSRC) under Project FleXEdge: Data-Driven Cloud-to-Edge Computing for Scalable Near Real-Time Local Flexibility Markets.}

\vspace{-4mm}
\section{Introduction}
The global power system decarbonization is pushing a higher integration of distributed energy resources (DERs), such as distributed renewables, energy storage (ES), and thermostatically controlled loads (TCLs). These DERs have valuable power flexibility, which is equivalent to that provided by multiple large power plants, with their global value estimated at USD~\$270~billion by 2040 \cite{iea_value}. However, the inherent uncertainties of DERs and their modelling complexity represent a significant challenge in unlocking their flexibility \cite{zhou2024aggregated}. Accurate DER modeling tends to be nonlinear and nonconvex. For example, the charging/discharging efficiency of ESS is observed to have a nonlinear nonconvex relationship to the charging/discharging power \cite{non-linear_efficiency}, and the accurate TCL model is a set of differential equations \cite{HVAC_modeling}. For computational tractability, traditional optimization methods often use convex or even linear approximations of these DER models, which can lead to suboptimal results in real-world operations. Reinforcement learning (RL) has emerged as a promising alternative that could learn to adapt to complex non-convex environments \cite{RL}. Initial attempts leverage traditional Q-learning for tasks like smart charging of electric vehicles \cite{RL_application1} and flexible load management \cite{RL_application2, RL_application3}, but are constrained by their inability to operate in continuous action spaces. Consequently, the focus has shifted towards more advanced policy-based algorithms, for instance, the application of Deep Deterministic Policy Gradient (DDPG) for smart home energy management \cite{RL_application4}, Proximal Policy Optimization (PPO) for microgrid energy management \cite{RL_application5}, and Soft Actor-Critic (SAC) for the optimal dispatch of wind-storage systems \cite{RL_application6}. However, most of the above studies employ standard RL algorithms trained from scratch, without a deep investigation into their intrinsic limitations, such as poor sample efficiency and the sub-optimality of the learned policy. Specifically, the difficulty and instability associated with training agents from scratch for complex systems present a major barrier to their real-world adoption.


The success of reinforcement learning from human feedback (RLHF) \cite{RLHF} in large language models (LLMs) provides valuable insights for DERs control. Standard LLM training follows the ``supervised pre-training plus RL fine-tuning'' paradigm, which consists of two stages: 1) pre-training through (self-)supervised learning, followed by 2) RLHF for fine-tuning \cite{RLHF+LLM}. This paradigm shift---using RL for fine-tuning rather than training from scratch---has been crucial for its successful real-world applications. The DERs' control problem shares key characteristics similar to LLM: both use RL for complicated problems with substantial data available for pre-training (there are sufficient historical DER operation experiences, though non-/sub-optimal or rule-based). This motivates us to migrate the two-stage paradigm of LLM training to DER control problems.

Some existing work has attempted to utilize demonstration data to train a DER control policy. Ref. \cite{IL1} develops a multi-agent imitation learning (IL) framework where individual agents are trained to emulate the optimal control actions of an optimization solver for managing multiple energy storage systems in a microgrid. Ref. \cite{IL2} proposes an IL-based approach that mimics an MILP solver for the economic dispatch of a microgrid. However, these studies are confined to supervised imitation and overlook the performance gains achievable through subsequent fine-tuning based on the interactive environment. Moreover, some work has tried to use historical DER operation data to enhance the performance of RL algorithms. For instance, Ref. \cite{application2} conducts offline convex optimizations on historical data, assuming perfect knowledge of all variables, to generate expert demonstrations for initializing the replay buffer. In contrast, rather than preparing demonstration data in a single batch, Ref. \cite{application1} establishes a continuous expert data flow by employing an optimizer to iteratively refine the trajectories collected by the RL agent. However, these studies primarily relies on offline RL methods, where prior data is simply stored in a replay buffer to facilitate training from scratch. 

Recent work has also explored multi-step training frameworks to enhance agents' performance for online deployment. For example, Ref. \cite{sim2real_1} proposes a two-stage method that first trains a calibrated and conservative value function from offline data and then fine-tunes the policy online to avoid performance degradation. Ref. \cite{sim2real_2} introduces an intermediate online training phase to train a new online value function, which then acts jointly with the frozen offline value function to refine the policy network. However, these studies focus on improving the estimation of value functions while overlooking the quality issues of the offline dataset. Suboptimal data can corrupt the value functions, causing the policy to overvalue inferior behaviors from the dataset and degrade the policy's adaptability to the online environment. Furthermore, these studies do not consider refining the policy in a high-fidelity simulated environment before online fine-tuning, a critical step for a smooth transition to the real-world environment with complex and stochastic dynamics.

Fundamentally, current methods do not follow the ``supervised pre-training plus RL fine-tuning'' paradigm of using prior data for pre-training and RL for fine-tuning. Given the success in LLM and the similarity in DER control problems, migrating the two-stage paradigm to DER control may lead to extra performance gain. Therefore, this work proposes the Supervised Reinforcement Learning (SRL) framework for DER control problems, involving: 1) a supervised pre-training stage for achieving baseline operational performance; 2) an RL-based fine-tuning stage for enhancing performance. This paper makes the following contributions:
\begin{enumerate}
    \item Motivated by the success in the current LLM training paradigm, we propose a SRL framework that combines supervised pre-training and RL-based fine-tuning for DER control problems. Specifically, a policy is pre-trained on demonstrations and then fine-tuned with RL methods. The proposed framework can effectively enhance the performance of the DER control policy.
    \item For the RL-based fine-tuning stage, we propose a two-step fine-tuning method, which involves an offline fine-tuning step in a simulated environment and an online fine-tuning step in the actual environment. This further improves the optimality of the control policy and provides a higher starting performance in the real environment.
    \item We performed comprehensive numerical experiments to validate the effectiveness and superiority of our proposed framework. Specifically, we evaluated its performance with varying quality of the pre-training demonstration data. Results show that performance improvement can be attained even with low-quality demonstration data.
\end{enumerate}

The rest of the paper is organized as follows: Section \ref{sec:problem} introduces the problem to be solved and the corresponding formulated Markov Decision Process (MDP). Section \ref{sec:method} presents the proposed SRL framework. Section \ref{sec:case} conducts case studies, while Section \ref{sec:conclusion} concludes the paper.

\section{Problem Formulation}
\label{sec:problem}
We consider an optimal energy management problem of a microgrid connected to an external grid. This microgrid integrates a heterogeneous collection of DERs, including photovoltaic (PV) panels, wind turbines (WTs), diesel generators (DGs), thermostatically controlled loads (TCLs), and energy storages (ESs). We assume a centralized control setting where a single controller governs the operation of all DERs. The main objective is to minimize the operation cost while satisfying the supply-demand balance.

This section first presents a complete and accurate mathematical modelling of the microgrid control problem. Subsequently, this problem is approximated as a Markov Decision Process (MDP) for the application of the RL paradigm.

\subsection{Accurate Mathematical Model}
The objective of DER control in the microgrid is to minimize the overall operational cost:
\begin{equation}\label{eq1}
\min \sum_{t\in T} \Big(\lambda_{\text{buy},t}P_{\text{buy},t} + \lambda_{\text{sell},t}P_{\text{sell},t} + \sum_{g\in \mathcal{G}}c_{g,t}P_{g,t} \Big),
\end{equation}
where $\lambda_{\text{buy},t}$ and $\lambda_{\text{sell},t}$ are the price of buying/selling energy from/to the external grid; $P_{\text{buy},t}$ and $P_{\text{sell},t}$ are the amount of energy that trade with the external grid; $P_{g,t}$ is the amount of energy generated by generator $g$ and $c_{g,t}$ is the corresponding generation cost (we assume a linear cost model). This paper focuses on accurate (nonlinear) DER operational models. The battery degradation cost and the wear-and-tear cost of TCLs are not considered for simplicity, but are straightforward to include. 

The first set of constraints in the microgrid DER operation is the power balance:
\begin{equation} 
    \begin{split}
         & \sum_{g\in \mathcal{G}, \mathcal{W}, \mathcal{P}}{P_{g,t}} + \sum_{m \in \mathcal{B}} P^{\text{dis}}_{m,t} + P^{\text{UPG}}_t  = P^D_t + \sum_{l \in \mathcal{H}}P_{l, t} + \sum_{m \in \mathcal{B}} P^{\text{ch}}_{m,t},  \\
         & \quad\quad\quad\quad\quad\quad\quad\quad\quad\quad\quad\quad\quad\quad\quad\quad\quad\quad\quad\quad\forall t\in T,
    \end{split}
\end{equation}
\begin{equation} 
P^{\text{UPG}}_t = P_{\text{buy},t} - P_{\text{sell},t} ,\forall t\in T,
\end{equation}
where $P^{\text{dis}/\text{ch}}_{m,t}$ denotes the discharging/charging power of ES $m$; $P^{\text{UPG}}_t$ is the power traded with the external grid; $P^{D}_t$ is local electricity demand of the microgrid; $P_{l, t}$ is the power demand of TCL $l$; $\mathcal{G}$ represent the sets of DGs; $\mathcal{W}$ represent the sets of WTs; $\mathcal{P}$ represent the sets of PVs; $\mathcal{H}$ represent the sets of TCLs; and $\mathcal{B}$ represent the sets of ESs.   

The accurate DER operation models represent the other part of the constraints of the microgrid DER operation problem:
\subsubsection{Distributed Energy Generator}
The distributed energy generators, including DGs, PVs, and WTs, can be modelled as:
\begin{equation} \label{eq2}
    \underline{P}_g \leq P_{g,t} \leq \overline{P}_g,\forall g\in \mathcal{G} \cup \mathcal{W} \cup \mathcal{P},\forall t\in T,
\end{equation}
where $\underline{P}_g$ and $\overline{P}_g$ are upper and lower boundaries for the output of generator $g$.

\subsubsection{Energy Storage}
An ES can be modelled as:
\begin{equation} \label{eq3}
    \begin{split}
    & S_{m,t+1} = S_{m,t} + \frac{P^{\text{ch}}_{m,t}\Delta{t}\eta^{\text{ch}}_m}{E_m} - \frac{P^{\text{dis}}_{m,t}\Delta{t}}{\eta^{\text{dis}}_m E_m}, \\
    &\qquad\qquad\qquad\qquad\qquad\forall m\in \mathcal{B},\forall t\in T ,
    \end{split}
\end{equation}
\begin{equation} \label{eq4}
    \underline{S}_{m} \leq S_{m,t} \leq \overline{S}_{m},\forall m\in \mathcal{B},\forall t\in T,
\end{equation}
\begin{equation} \label{eq5}
    0 \leq P^{\text{ch}}_{m,t} \leq \overline{P}_{m}\mathcal{I}_{m,t},\forall m\in \mathcal{B},\forall t\in T,
\end{equation}

\begin{equation} \label{eq6}
0 \leq P^{\text{dis}}_{m,t} \leq \overline{P}_{m}\left(1-\mathcal{I}_{m,t}\right),\forall m\in \mathcal{B},\forall t\in T,
\end{equation}
where Eq. \eqref{eq3} represents the State of Charge (SoC) dynamics of an ES with energy capacity $E^{\text{es}}_m$, considering the charging efficiency $\eta^{\text{esc}}_m$ and discharging efficiency $\eta^{\text{esd}}_m$. Eq. \eqref{eq4} indicates the limited boundary for ESs' SoC. Eq. \eqref{eq5} and \eqref{eq6} demonstrate the upper and lower bounds of charging and discharging power with a binary variable $\mathcal{I}_{m,t}$ distinguishing the status of the ES between charging ($\mathcal{I}_{m,t} = 1$) and discharging ($\mathcal{I}_{m,t} = 0$).

Although the charging/discharging efficiency is usually assumed to be constant in the conventional ES model, the efficiency is actually a nonlinear function of SoC and charging/discharging power \cite{non-linear_efficiency}. To account for the nonlinear relationship among charging efficiency, SoC, and charging power, we adopt the steady state equivalent circuit model proposed in \cite{ES_model}, in which the circuit parameters of ES $m$ can be formulated as:
\begin{equation} \label{eq7}
    V^{\text{oc}}_m=a_0 e^{\left(-a_1 S_m\right)}+a_2+a_3 S_m-a_4 S_m^2+a_5 S_m^3,
\end{equation}
\begin{equation} \label{eq8}
    R^s_m=b_0 e^{\left(-b_1 S_m\right)}+b_2+b_3 S_m-b_4 S_m^2+b_5 S_m^3,
\end{equation}
\begin{equation} \label{eq9}
    R^{\text{ts}}_m=c_0 \cdot e^{-c_1 \cdot S_m}+c_2,
\end{equation}
\begin{equation} \label{eq10}
    R^{\text{tl}}_m=d_0 \cdot e^{-d_1 \cdot S_m}+d_2,
\end{equation}
where $V^{\text{oc}}_m$ is open circuit voltage; $R^s_m$ is the series resistance that used to characterize the charge/discharge energy losses; $R^{\text{ts}}_m$ and $R^{\text{tl}}_m$ are resistances used to characterize the short-term and long-term transient responses of the ES.

For a particular SoC and output power, the circuit current can be obtained by solving the quadratic equation $P^e_m=I_m\left[V^{\text{oc}}_m-\left(R^s_m+R^{\text{ts}}_m+R^{\text{tl}}_m\right) I_m\right]$.
\begin{equation} \label{eq11}
    I_m=\frac{V^{\text{oc}}_m-\sqrt{\left(V^{\text{oc}}_m\right)^2-4 \cdot \left(R^s_m+R^{\text{ts}}_m+R^{\text{tl}}_m\right) \cdot P^{\text{ch}/\text{dis}}_m}}{2 \cdot \left(R^s_m+R^{\text{ts}}_m+R^{\text{tl}}_m\right)},
\end{equation}
where $I_m$ is the circuit current.

Subsequently, the charging and discharging efficiency of the ES is given by:
\begin{equation} \label{eq12}
    \eta^{\text{ch}}_m =\frac{V^{\text{oc}}_m}{V^{\text{oc}}_m-\left(R^s_m+R^{\text{ts}}_m+R^{\text{tl}}_m\right) \cdot I_m},
\end{equation}
\begin{equation} \label{eq13}
    \eta^{\text{dis}}_m =\frac{V^{\text{oc}}_m-\left(R^s_m+R^{\text{ts}}_m+R^{\text{tl}}_m\right) \cdot I_m}{V^{\text{oc}}_m}.
\end{equation}

\subsubsection{Thermostatically Controlled Load}
The operation and the temperature evolution of a TCL are modeled through the following first-order differentiation equations \cite{TCL_model}:
\begin{equation} \label{eq14}
    0 \leq P_{l,t} \leq \overline{P}_l,\forall l\in \mathcal{H},\ \forall t\in T,
\end{equation}
\begin{equation} \label{eq15}
    \begin{split}
    & \dot{\theta}_t=-\frac{1}{R_{\text{th}} C_{\text{th}}}\left(\theta_t-\theta^a_t\right)-\frac{\boldsymbol{\kappa}}{C_{\text{th}}} P_{l,t},\ \forall l\in \mathcal{H},\forall t\in T,
    \end{split}
\end{equation}
\begin{equation} \label{eq16}
\theta_t \in [\theta^{\text{set}} - \Delta\theta, \theta^{\text{set}} + \Delta\theta]
\end{equation}
where $\overline{P}_l$ is rated power of TCL $l$; $\theta_t$, $\theta^a_t$, and $\dot{\theta}_t$ are the indoor temperature, ambient temperature, temperature set-point, and change of temperature, respectively; $R_{\text{th}}$, $C_{\text{th}}$, and $\boldsymbol{\kappa}$ are thermal resistance, thermal capacitance, and coefficient of performance; $\theta^{\text{set}}$ is the temperature set-point and $[\theta^{\text{set}} - \Delta\theta, \theta^{\text{set}} + \Delta\theta]$ represents the confort zone.

It is worth noting that the proposed framework does not depend on specific DER operation models. SRL, as a model-free method, is readily compatible with non-differentiable or even black-box models, such as any battery vendor simulators. The accurate modeling presented in this subsection is primarily for constructing a high-fidelity simulation environment for validation, aiming to approximate the real-world environment.

\subsection{MDP Formulation of DER Operation}
To facilitate the use of RL, we can approximate the DER operation problem as an MDP, which is characterized by a tuple $\left<\mathcal{S},\mathcal{A},\mathcal{P},r,\gamma\right>$, where $\mathcal{S}$ is the state space; $\mathcal{A}$ is the action space; $\mathcal{P}\left(s'\lvert s, a\right):\mathcal{S}\times\mathcal{A}\times\mathcal{S}\rightarrow \mathbb{R}$ is the probability of state transition; $r: \mathcal{S} \times \mathcal{A} \rightarrow \mathbb{R}$ is the reward function; $\gamma \in [0,1]$ is the discount factor for future reward.

\subsubsection{State}
The state space at time $t$ can be defined as:
\begin{equation} \label{eq16}
    \mathbf{s}_t = [\mathbf{S}_t, \mathbf{P}^{\text{pv}}_t, \mathbf{P}^{\text{wt}}_t, \mathbf{T}^{\text{in}}_t, \mathbf{T}^{a}_t, P^{D}_t,    \lambda^{p}_{\text{buy},t},\lambda^{p}_{\text{sell},t}, t],
\end{equation}
where $\mathbf{S}_t$ is the SoC of ESs; $\mathbf{P}^{\text{pv}}_t$ and $\mathbf{P}^{\text{wt}}_t$ are renewable energy output; $\mathbf{T}^{\text{in}}_t$ is indoor temperature of the TCL; $\mathbf{T}^{a}_t$ is the ambient temperature; $\lambda^{p}_{\text{buy},t}$ and $\lambda^{p}_{\text{sell},t}$ are the buying and selling prices of energy from the external grid at the current time step.

\subsubsection{Action}
The action space at time $t$ is defined as:
\begin{equation}
    \mathbf{a}_t = [\boldsymbol{a}^{\text{dg}}_t, \boldsymbol{a}^{\text{es}}_t, \boldsymbol{a}^{\text{tcl}}_t],
\end{equation}
where $\boldsymbol{a}^{\text{dg}}_t$ and $\boldsymbol{a}^{\text{es}}_t$ are the energy output rate for DGs and ESs; and $\boldsymbol{a}^{\text{tcl}}_t$ is the energy input rate of TCL. It should be noted that the actions for DG and TCL are normalized rates within $[0, 1]$, and the action for ES is a charge/discharge rate within $[-1, 1]$. The actual power output is obtained by scaling these actions with the device's maximum power capacity, thus inherently guaranteeing compliance with physical limits.

\subsubsection{State Transition}
The state transition from $\mathbf{s}_t$ to $\mathbf{s}_{t+1}$ is characterized by both deterministic and stochastic elements. We can decompose the state vector $\mathbf{s}_t$ into two components: endogenous states, which are directly controllable, and exogenous states, which are influenced by the external environment. The endogenous components, such as the SoC of ESs $\mathbf{S}_t$ and TCL indoor temperatures $\mathbf{T}^{\text{in}}_t$, follow a deterministic transition function dependent on the current state and the applied action $\mathbf{a}_t$. Conversely, the exogenous components, representing renewable generation, energy prices, and ambient conditions, evolve stochastically, driven by unobservable environmental randomness.

\subsubsection{Reward}
The reward function $r_t$ is formulated as the negative of the total operational cost. To this end, the agent can learn a policy that minimizes long-term costs by maximizing the expected cumulative reward.
\begin{equation} \label{eq18}
    r_t = -\Big(\lambda_{\text{buy},t}P_{\text{buy},t} + \lambda_{\text{sell},t}P_{\text{sell},t} + \sum_{g\in \mathcal{G}}c_{g,t}P_{g,t}\Big).
\end{equation}

\section{Methodology}
\label{sec:method}
\begin{figure}[t]
\centering
\includegraphics[width=0.40\textwidth]{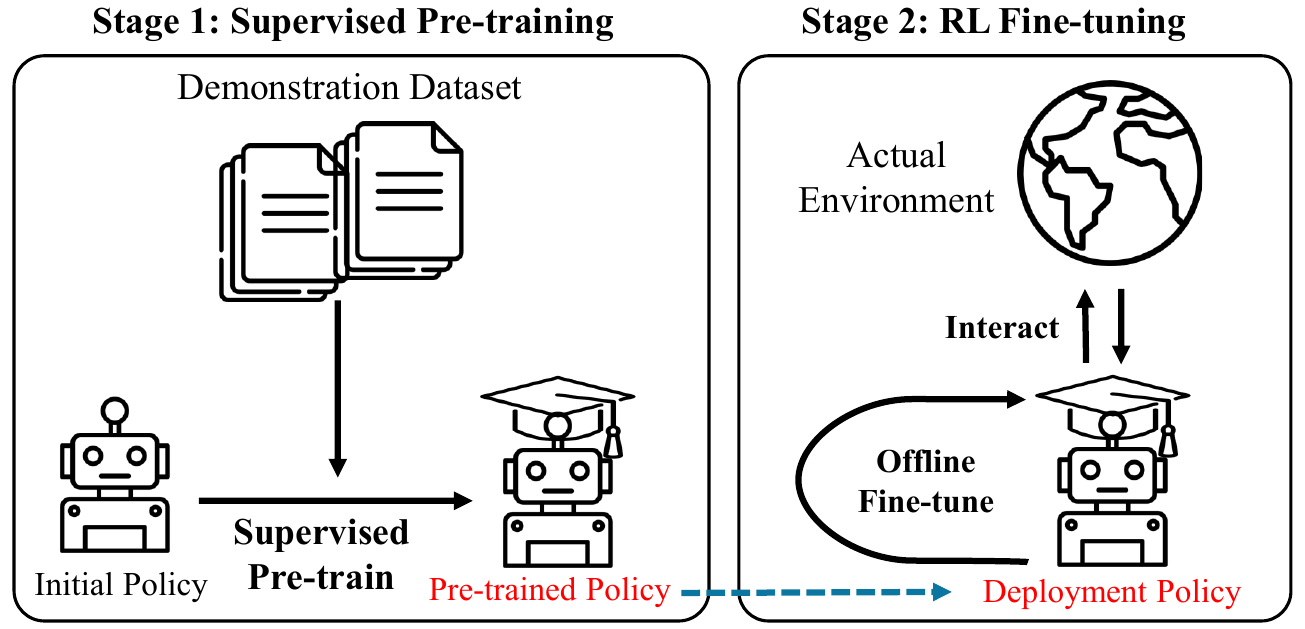}
\caption{Schematic of the proposed SRL framework.}
\label{fig_framework}
\vspace{-12pt}
\end{figure}

The proposed SRL framework for DER control is depicted in Fig. \ref{fig_framework}. The framework begins with a supervised pre-training stage to establish a pre-trained policy that realizes a baseline level of operational performance. This is followed by a fine-tuning stage, where the policy is further refined to improve its efficacy and adapt to the real-world environment.

\subsection{Supervised Pre-training From Demonstrations}
The supervised pre-training stage serves as a crucial foundation for our framework, motivated by the need to enhance sample efficiency and ensure an improved starting point for learning. Drawing inspiration from the two-stage paradigm of LLM training, this stage leverages demonstration data to initialize the control policy with a competent behavioral prior. To be more specific, we pre-train a policy $\mu$ based on the demonstration dataset $\dot{\mathcal{D}}$ with previously collected data in a supervised learning manner. Such prior data may come either from look-back optimization problems based on historical information or from human expert operations used in real production environments. The loss function of the pre-training stage can be the standard squared loss in supervised regression tasks: 
\begin{equation}
\label{eq_il}
    L(\boldsymbol{\theta}) = \mathbb{E}_{(\boldsymbol{s},\boldsymbol{a}) \sim \dot{\mathcal{D}}} \left[ \frac{1}{2} \left( \mu(\boldsymbol{s} ; \boldsymbol{\theta})-\boldsymbol{a} \right)^2 \right],
\end{equation}
where $\boldsymbol{\theta}$ is the parameter of the policy, $(\boldsymbol{s},\boldsymbol{a})$ denotes state-action pairs that are previously collected. By minimizing the error between the demonstration action and the policy output under a certain state, the agent learns to predict the most probable action corresponding to a given state.

This supervised pre-training stage avoids the critical disadvantage of traditional RL, which often starts from scratch and relies on inefficient, random exploration that can be slow and unstable. Instead of exploring blindly, our pre-trained agent begins its reinforcement learning process with an inherent understanding of competent behavior, enabling it to generate meaningful and effective transitions from the outset. This ``warm start" not only dramatically improves sample efficiency but also provides a solid foundation for the subsequent fine-tuning stage.

\subsection{Two-step Fine-tuning}
\label{sec:SRL2}


\begin{figure}[t]
\centering
\includegraphics[width=0.48\textwidth]{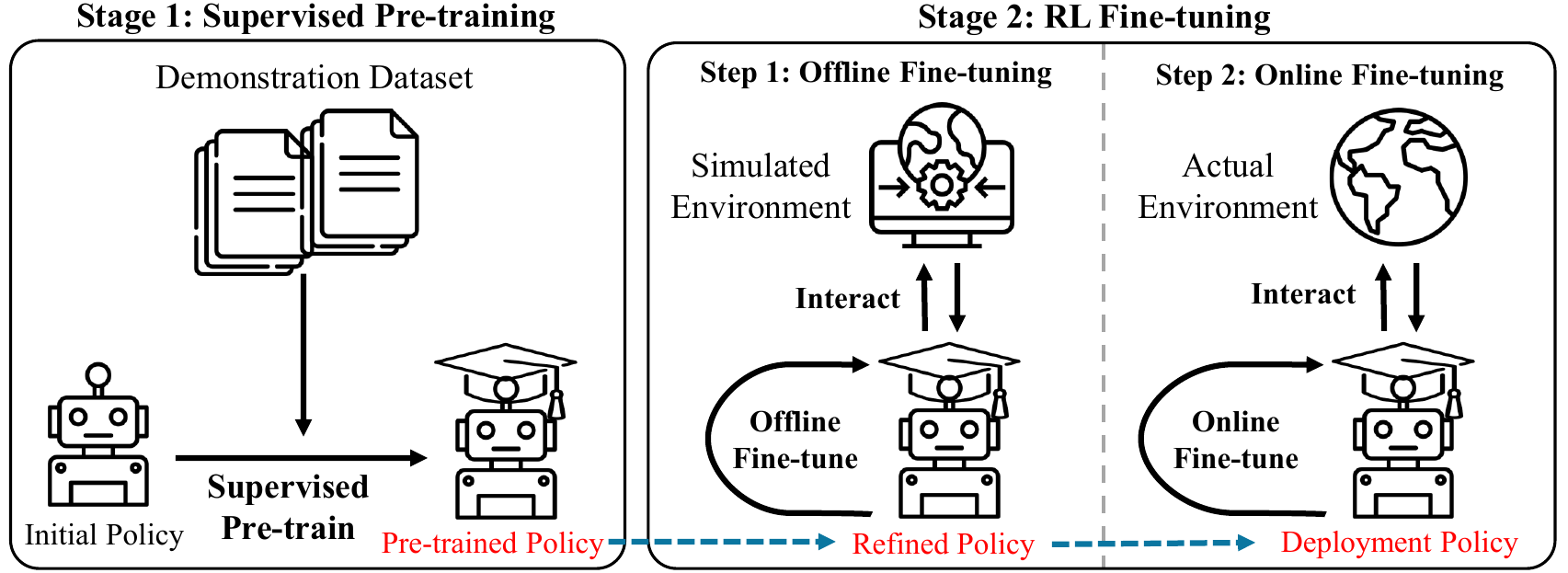}
\caption{The SRL framework with two-step fine-tuning.}
\label{fig_framework2}
\vspace{-12pt}
\end{figure}

The motivation for adopting the fine-tuning procedure stems from the inherent limitations of a policy derived purely from supervised learning. While pre-training establishes a competent baseline, the policy is essentially a static imitation of the demonstration dataset; i.e., the policy learned by minimizing the MSE loss does not necessarily lead to the lowest operational cost (highest reward). Therefore, a fine-tuning stage is required to allow the agent to adapt to the environment's dynamics and refine its strategy based on the received reward signal, thereby improving its performance on the control task. 

However, deploying the pre-trained policy directly into a real-world environment with its complex and stochastic dynamics poses a significant risk of the agent converging to a suboptimal solution. To mitigate the aforementioned risk and systematically bridge the policy from imitation to optimization, we split the fine-tuning stage into two steps: offline fine-tuning in the simulated environment and online fine-tuning in the actual environment, as shown in Fig. \ref{fig_framework2}. Through offline fine-tuning in the simulated environment, the agent can cultivate a preliminary decision-making logic oriented towards maximizing long-term rewards. Then, the online fine-tuning step has a distinct and focused objective: closing the `sim-to-real' gap. This final step adapts the well-prepared policy to the unmodeled dynamics of the real-world physical system, ensuring truly optimal and adaptive performance in the actual operational environment.

A simulated environment resembles the actual complex environment, but with some approximation error. To model this approximation error, when creating the simulated environment we perform piecewise linearization to approximate the actual nonlinear dependency of ES efficiency on the SOC and charging/discharging power. Specifically, a fixed efficiency value is assigned to each discretized bin of SOC and ES power, indexed by $i$ and $j$:
\begin{equation} \label{eq_discretize1}
    \begin{split}
        & \eta^{\text{ch}/\text{dis}}_m (S_m, P^{\text{ch}/\text{dis}}_m)= \eta^{\text{ch}/\text{dis}}_{i,j}, \\
        & \forall S_m \in\left[S O C_i, S O C_{i+1}\right], P^{\text{ch}/\text{dis}}_m \in\left[P_j, P_{j+1}\right].
    \end{split}
\end{equation}

As for the differential process of temperature evolution, the continuous linear differential process governing temperature evolution is discretized using a forward Euler method. Specifically, the differential Eq. \eqref{eq15} is transformed into a discrete-time difference equation using a fixed time step $\Delta t$:
\begin{equation} \label{eq_discretize2}
    \begin{split}
    & \theta_{k+1}=\theta_{k} - \Delta t\left[ \frac{1}{R_{\text{th}} C_{\text{th}}}\left(\theta_{k} - \theta^a_{k}\right)+\frac{\boldsymbol{\kappa}}{C_{\text{th}}} P_{l,k}  \right].
    \end{split}
\end{equation}

It should be noted that our proposed framework is designed to be compatible with a wide range of policy-based RL methods. For our implementation, we selected Proximal Policy Optimization (PPO) \cite{PPO} as the foundational algorithm, primarily due to its renowned stability and consistent performance.

\subsection{Full Algorithm}
\begin{algorithm}[t]
	\caption{Supervised Reinforcement Learning}
	\label{full_algo} 
	\begin{algorithmic}[1]
		\STATE Initialize policy network $\pi_{\boldsymbol{\theta}}$ and replay buffer $\mathcal{D}\leftarrow\emptyset$
        \STATE Collect demonstration data for supervised pre-training
        \FOR{episode $s=1$ to $E_1$}
            \STATE Update policy weights $\boldsymbol{\theta}$ according to Eq. (\ref{eq_il})
        \ENDFOR
        \STATE Create simulated environment according to Eqs. (\ref{eq_discretize1}) and (\ref{eq_discretize2}) for offline fine-tuning
		\FOR{episode $s=1$ to $E_2$}
            \STATE Sample action $\mathbf{a}_t$ according to $\pi_{\boldsymbol{\theta}}$ 
            \STATE Observe the next state $\mathbf{s}_{t+1}$ and reward $r_t$
            \STATE Store transition $\{\mathbf{s}_{t}, \mathbf{a}_t, \mathbf{s}_{t+1}, r_t\}$ to the replay buffer $\mathcal{D}$
            \STATE Sample a batch of experience from the replay buffer $\mathcal{D}$
            \STATE Update policy weights $\boldsymbol{\theta}$ based on the PPO paradigm
        \ENDFOR
        \STATE Initialize replay buffer $\mathcal{D}\leftarrow\emptyset$ and prepare real-world environment for online fine-tuning
        \FOR{episode $s=1$ to $E_3$}
            \STATE Sample action $\mathbf{a}_t$ according to $\pi_{\boldsymbol{\theta}}$
            \STATE Observe the next state $\mathbf{s}_{t+1}$ and reward $r_t$
            \STATE Store transition $\{\mathbf{s}_{t}, \mathbf{a}_t, \mathbf{s}_{t+1}, r_t\}$ to the replay buffer $\mathcal{D}$
            \STATE Sample a batch of experience from the replay buffer $\mathcal{D}$
            \STATE Update policy weights $\boldsymbol{\theta}$ based on the PPO paradigm
        \ENDFOR
	\end{algorithmic} 
\end{algorithm}

The complete procedure of the proposed SRL framework is detailed in Algorithm \ref{full_algo}. The algorithm is structured into two principal stages: a supervised pre-training stage to establish a baseline policy, followed by a two-step fine-tuning stage encompassing both offline and online refinement to enhance performance and real-world adaptability.

The algorithm commences with supervised pre-training (Lines 1-5), where the policy is trained on a demonstration dataset by minimizing the loss function Eq. (\ref{eq_il}). This stage establishes a behavioral prior before any environmental interaction, which is followed by offline fine-tuning (Lines 6-13) in a simulated environment. The policy is further refined through interaction, using the PPO algorithm to update its weights. The final step is online fine-tuning (Lines 14-21). Here, the refined policy is deployed in the actual environment to bridge the sim-to-real gap.

\section{Case Studies}
\label{sec:case}

\subsection{Experiment Setups}
For the DERs control problem, we consider 1 PV, 1 WT, 1 DG, 1 ES, and 1 TCL. The operational data of power demand and renewable energy generation are sourced from the UK Nation Grid \cite{DATASET1}. The ambient temperature data is obtained from \cite{Temp_dataset}. In this work, we adopt a time-of-use (TOU) electricity pricing scheme modified from \cite{ToU}, with details presented in Table \ref{tab_Elec_price}. The price of selling power is set to 50\% of the buying price. The DG cost coefficient $c_{g,t}$ is set to 0.2 \pounds/kWh. The initial SoC of ESs is set to be 0. The constructed dataset spans a continuous period of five months. For our proposed framework, we allocate the first month as the pre-training set, the second month as the offline fine-tuning set, the third and fourth months as the online fine-tuning set, and the final month as the test set. For the standard RL method, the first four months would be allocated as the training set, and the last month as the test set.

\begin{table}[t]
\caption{Time-of-Use Pricing Scheme}
\centering
\renewcommand\arraystretch{1.15}
\begin{tabular}{ccc}
\toprule
Type & Time Slot & \begin{tabular}[c]{@{}c@{}}Buying Price\\ ($\pounds$/kWh)\end{tabular} \\ \hline
Off-Peak & 23:00 - 7:00               & 0.04  \\ 
Peak   & 7:00 - 10:00; 16:00 - 21:00  & 0.30 \\
Normal & 10:00 - 16:00; 21:00 - 23:00 & 0.12 \\
\bottomrule
\end{tabular}
\label{tab_Elec_price}
\end{table}

To validate the performance, this paper evaluates the following six methods, including two methods under the proposed SRL framework and four benchmark methods.

\subsubsection{Vanilla Supervised Reinforcement Learning (SRL-1)} This is the vanilla implementation of the proposed SRL framework, where the fine-tuning stage only consists of a single online fine-tuning step.

\subsubsection{Supervised Reinforcement Learning with Two-step Fine-tuning (SRL-2)} This is the proposed SRL framework with the proposed two-step fine-tuning method as illustrated in Section \ref{sec:SRL2}.

\subsubsection{Pre-trained Policy (PP)} A policy is only pretrained with demonstrations in a supervised manner, which represents the baseline performance that is achieved before fine-tuning. 

\subsubsection{Vanilla PPO} This benchmark implements a standard PPO agent trained entirely from scratch. This baseline agent's policy is randomly initialized and refined exclusively through online interaction with the environment.

\subsubsection{Stochastic Model Predictive Control (sMPC)} To handle uncertainty, this method extends standard MPC by reformulating the objective in expectation form. The expectation is estimated using a scenario-based approach. First, a large pool of 5,000 scenarios of the random variables is generated, and the scenario reduction technique in \cite{Scen_reduction} is applied to distill the 10 most representative scenarios. Similar to standard MPC, sMPC solves an optimisation problem with a given look-ahead horizon but only implements the first-step solution. This process is then repeated at each decision step.

\subsubsection{Perfect MILP} To establish a performance upper bound, a deterministic MILP benchmark is formulated under the ideal condition of perfect information, i.e., the accurate and nonconvex system model and all future uncertainties are precisely known.

To simulate forecasting uncertainty, we modeled all relevant inputs using normal distributions. The ground truth value served as the mean, with a standard deviation set to 15\% of the mean for demand and renewables, and 5\% for temperature. This approach was used consistently to generate scenarios for sMPC and to create the training and test sets for data-driven methods. All experiments are carried out on a workstation with NVIDIA GeForce RTX 3080 Ti GPU, Intel(R) Xeon(R) W-3335 CPU, and 188GB RAM. All optimization problems were solved using the Gurobi Optimizer (version 10.0.3). All codes are written based on the Python language and PyTorch framework.

\subsection{Demonstration Generation}

The demonstration data for pre-training can be derived from two primary sources: (1) historical logs of human decisions, which are often suboptimal due to uncertainty at the time of decision-making; and (2) trajectories generated by re-solving the problem using simulation models after all uncertain parameters have been revealed. In the latter case, the problem may not be solved to global optimality due to limited computational resources or time constraints, and the simulation model may be inaccurate or biased, both of which can result in suboptimal demonstration data.

To mimic these imperfections in the real-world data sources, our case study employs a demonstration data generation method. Leveraging the state distribution from our pre-training dataset and the constructed simulation model, we use a deterministic optimization solver to find the control actions for each given state. This procedure results in a collection of state-action pairs $(\boldsymbol{s},\boldsymbol{a})$, which form the basis of our demonstration dataset $\dot{\mathcal{D}}$. To mimic the imperfections, we introduce an additional equality constraint to the optimization problem:
\begin{equation}\label{eq:quality}
    O = \rho \, \Tilde{O},
\end{equation}
where $O$ is the objective function \eqref{eq1} of the optimization problem of the simulated models; $\rho$ is the coefficient that controls the data quality; $\Tilde{O}$ is theoretical optimum (solved without the quality constraint \eqref{eq:quality}). This constraint intentionally relaxes the solution's optimality by forcing the objective value to be a specific multiple (e.g., 1.1, 1.2, or 1.3) of the optimal objective values found without the constraint. The distinct datasets embed different levels of control expertise:
\begin{itemize}
    \item Near-Optimal (NO) Data: Without the quality constraint, the obtained dataset represents a highly proficient expert policy within a simulated environment, in which the embedded knowledge reflects the efficient utilization of both DG and ES. When applied to the real-world target, this can serve as a near-optimal benchmark.
    \item High-Quality (HQ) Data: By setting $\rho = 1.1$, the dataset contains suboptimal but still highly competent actions. The control logic demonstrates a rational and effective utilization of both DG and ES resources.
    \item Medium-to-High-Quality (MtHQ) Data: By setting $\rho = 1.2$, the embedded expertise in the dataset further degrades. The policy reflects a reasonable dispatch strategy for DG but fails to use ES.
    \item Low-Quality (LQ) Data: By setting $\rho = 1.3$, the dataset represents a significantly flawed expert. The rationality of DG utilization is further diminished, occasionally including inefficient or counterintuitive actions, such as using DG in a low-price period, while the ES remains entirely unutilized.
\end{itemize}

It should be noted that this work does not explicitly address rare extreme events. The online fine-tuning step of SRL theoretically allows the agent to adapt, but this reactive process may still incur extra costs. A proactive approach would be to integrate extreme scenarios into both the demonstration data and the simulated environment, and to include them as part of the performance evaluation. Also note that the demonstration data generation process is agnostic to state distributions.


\subsection{Training Performance of SRL-1}
\begin{figure}[t]
\centering
\includegraphics[width=0.45\textwidth]{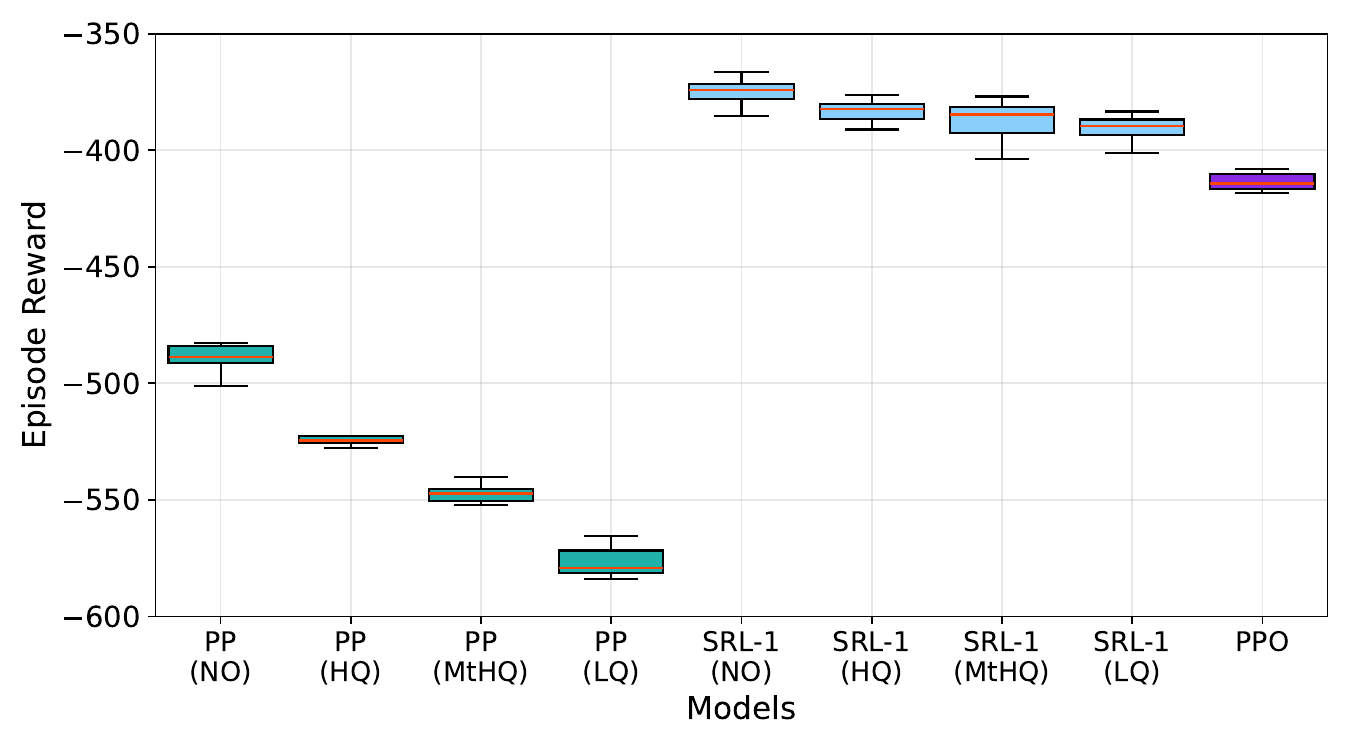}
\caption{Comparison of algorithms' convergence performance with varying levels of quality for demonstration data. The result displays the distribution of episode reward for each algorithm, aggregated over 10 independent trials.}
\label{fig_training1}
\end{figure}

We first validate the effectiveness of our proposed framework without the offline fine-tuning step in the context of DER control and investigate its robustness to varying demonstration data quality. The results of the experiment are presented in Fig. \ref{fig_training1}, which shows the convergent episode rewards for each algorithm in ten independent runs. The suffixes appended to the algorithm names denote the quality of the demonstration data used for pre-training.

First, SRL-1 consistently outperforms PP and PPO across all levels of demonstration quality, validating the effectiveness of the proposed SRL framework. The results also demonstrate the notable robustness of SRL-1 to data quality: even as demonstration quality decreases from near-optimal to low, SRL-1 still converges to a policy that outperforms the standard PPO baseline, highlighting its ability to effectively leverage and refine even imperfect expert knowledge. As a side note, when comparing PP and PPO, it is interesting to observe that pure supervised learning (even with the highest-quality demonstrations) still underperforms RL trained from scratch. This suggests a slight misalignment between the supervised loss and the objective of training a control policy, and underscores the superior alignment of RL methods in control problems.


\subsection{Ablation Study of Offline Fine-tuning}
\begin{figure}[t]
\centering
\includegraphics[width=0.45\textwidth]{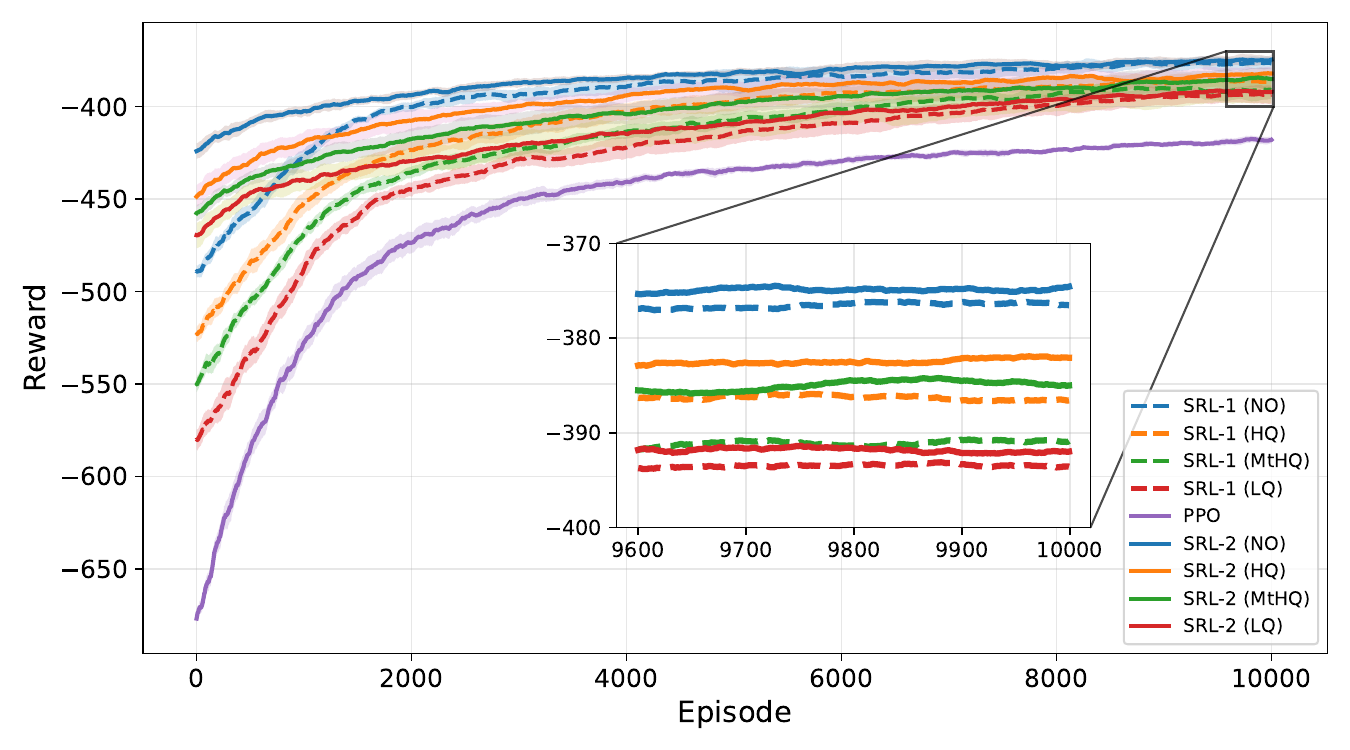}
\caption{The online fine-tuning reward curves of SRL-1 and SRL-2 with varying levels of demonstration quality. Note that SRL methods have gone through an offline fine-tuning step.}
\label{fig_training2}
\end{figure}
This section provides an ablation study to verify the effectiveness of the proposed two-step fine-tuning method in Section \ref{sec:SRL2}. It should be noted that the duration of the offline fine-tuning is a critical hyperparameter. Too few episodes may leave the policy unprepared, while too many risk overfitting to the simulation, which could impede sim-to-real transfer during the final online fine-tuning. Based on preliminary tests, we established a range of 2000-4000 episodes as effective. For this experiment, we fine-tuned the pre-trained policy for 3000 episodes in the simulated environment.

The results are presented in Fig.~\ref{fig_training2}, which shows the average training reward curves (in the online fine-tuning step) of ten independent runs for each method. It can be observed that, for any given quality of demonstration data, SRL-2 outperforms SRL-1 in both the initial performance at the start of online training and the final converged reward. A potential reason is that the offline fine-tuning step makes SRL-2 better aligned to the control objective. Consequently, when transferred to the real environment, although the sim-to-real gap still exists, the fine-tuned policy can immediately explore higher-quality experiences, making it less likely to be trapped in local optima and enabling it to achieve a higher reward upon convergence.
Overall, the results confirm that the offline fine-tuning is a critical component for enhancing real-world performance, particularly when leveraging realistically imperfect demonstration data for the pre-training.

\subsection{Operation Analysis}
\begin{figure}[t]
\centering
\includegraphics[width=0.45\textwidth]{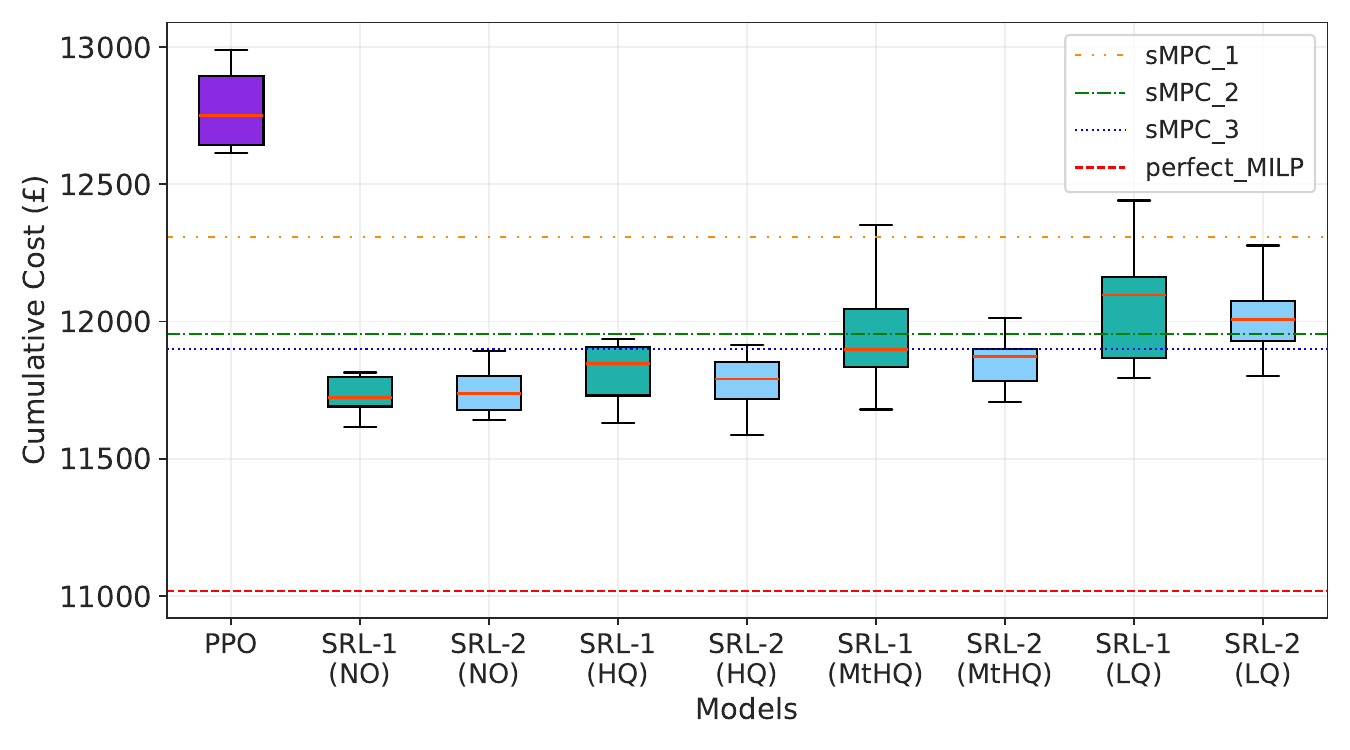}
\caption{Comparison of cumulative cost of the 30-day test period. Results are aggregated from 10 independent experimental runs.}
\label{fig_test}
\vspace{-12pt}
\end{figure}

To evaluate generalization performance, we tested the proposed framework against the benchmarks on a dedicated test set comprising 30 days of unseen data. It is important to note that for the sMPC benchmark, we limited the comparison to prediction horizons of 1 to 3 hours. This setting was based on two key factors: information advantage and computational tractability. Firstly, our RL paradigm operates on single-step predictive information, and providing sMPC with an extended multi-hour forecast would constitute an unfair informational advantage. Secondly, the sMPC algorithm's reliance on scenario generation leads to a significant increase in computational complexity as the prediction horizon lengthens, compromising its real-time feasibility. The typical solve times for 30-day test case of perfect\_MILP, sMPC\_1, sMPC\_2, sMPC\_3 are 6.95s, 3666.72s, 4351.48s, 5030.21s, respectively.

The cumulative operational costs for all algorithms over the 30-day test period are presented in Fig. \ref{fig_test}. The PPO agent, trained from scratch without the benefit of demonstration guidance or model-based information, performs the worst (highest cost). sMPC with a 1-hour horizon (sMPC-1) performs the second worst as the short timeframe is insufficient for efficient energy storage scheduling. Notably, the SRL-1, when pre-trained on HQ demonstrations, surpasses the performance of sMPC-3, even though the SRL-1 has information disadvantages. More impressively, our proposed SRL-2 surpasses sMPC-3 with only MtHQ demonstration data. The top-performing model, SRL-2 pre-trained with NO demonstrations, demonstrates the closest performance to the theoretical optimum (perfect MILP), with an average cumulative cost only \pounds 723.88 higher, corresponding to a minimal performance gap of 6.57\%.

\begin{figure}[t]
\centering
\includegraphics[width=0.50\textwidth]{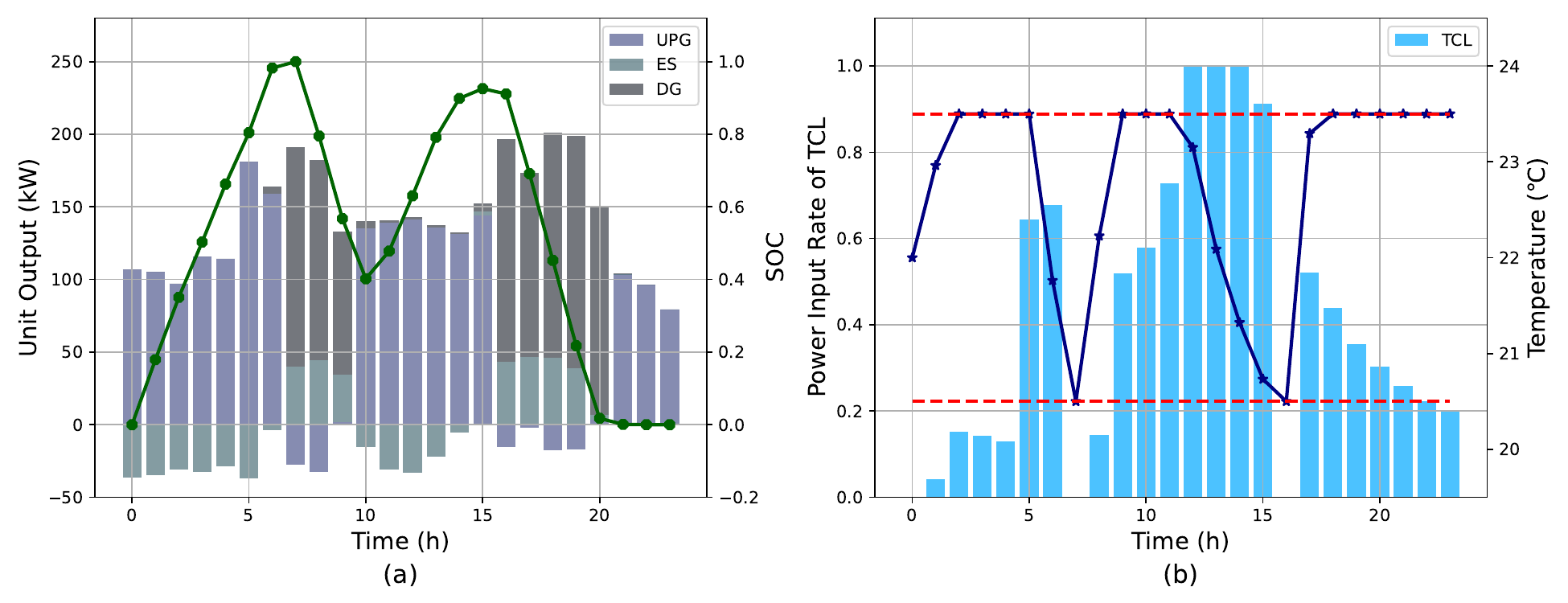}
\caption{Average behavior of the proposed algorithm for controlling DERs over the 30 test days: (a) energy dispatch result and SoC of ES; (b) TCL dispatch result and evolution of indoor temperature}
\label{fig_behavior}
\vspace{-12pt}
\end{figure}

In addition to quantitative evaluation, a qualitative sanity check was performed by visualizing the control behavior of the SRL-2 trained with HQ demonstrations on the test set, with the results presented in Fig. \ref{fig_behavior}. This analysis confirms that the agent learned an intuitive and economically rational policy. Fig. \ref{fig_behavior}(a) illustrates the average daily dispatch behavior of the energy units and the battery's SOC. The agent's strategy clearly adapts to electricity prices: during low-price periods, local demand is met by the external grid, and the ES is charged. In contrast, during high-price periods, the agent switches to local power generation via the DG and discharges the ES to minimize costs. Fig. \ref{fig_behavior}(b) details the agent's control of the TCL, showing the average indoor temperature relative to the temperature dead-band (indicated by the red dashed envelope). Prior to the onset of the two daily price peaks, the agent increases the TCL's energy input rate to pre-cool the space. By lowering the temperature within the comfort band ahead of time, the agent effectively reduces the need for costly cooling during peak-price hours, thereby optimizing the overall electricity bill.

\subsection{Scalability Analysis}
In this section, a scalability test is conducted to verify the effectiveness of the proposed SRL on DER coordination problems of varying scales. We conducted experiments with different numbers of DERs, ranging from 5 to 120. The proposed framework is compared against the standard PPO baseline in terms of both training time and test performance.

\begin{figure}[t]
\centering
\includegraphics[width=0.50\textwidth]{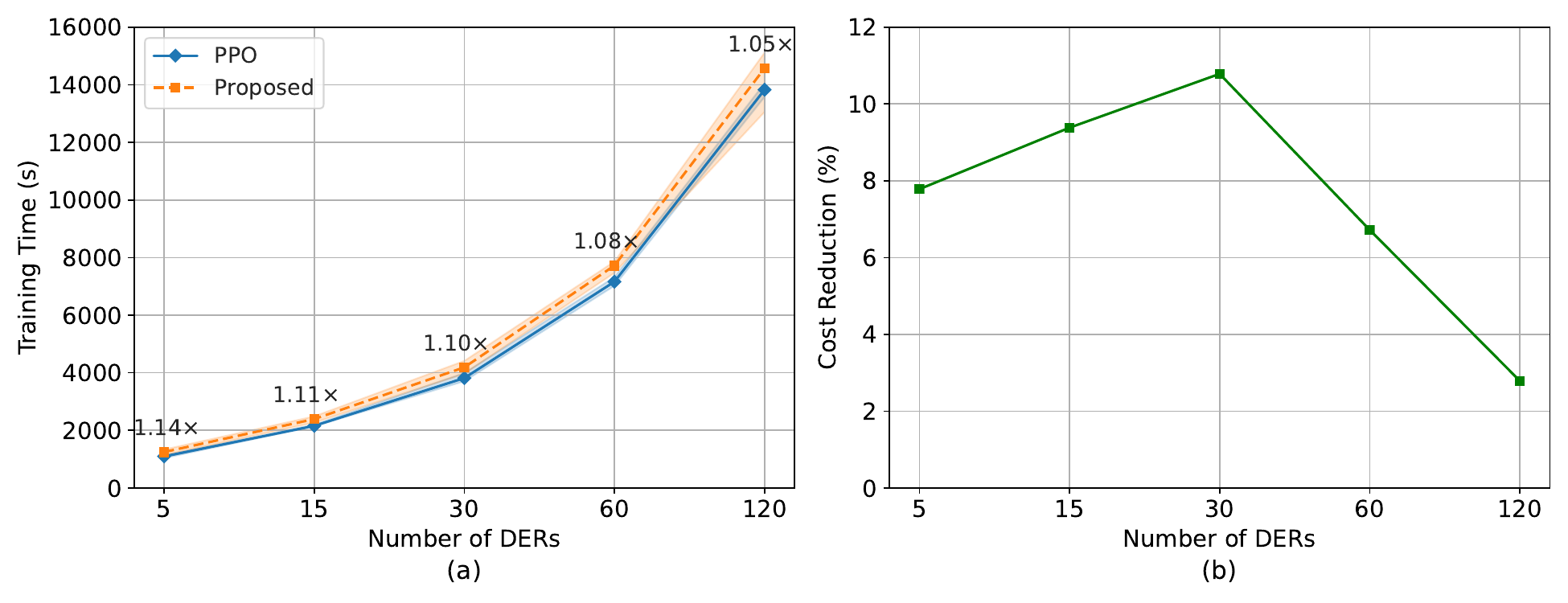}
\caption{Scalability test of the proposed SRL for different numbers of DERs, aggregated over 10 independent trials: (a) training time compared with standard PPO; (b) cost reduction compared with standard PPO}
\label{fig_scalability}
\vspace{-12pt}
\end{figure}

The comparison of training times between the proposed SRL and the standard PPO is presented in Fig. \ref{fig_scalability}(a). As depicted, the training duration for both methods escalates with the number of DERs. The case of 120 DERs requires nearly 12 times the training time of the case of 5 DERs. This is because the growth in DERs leads to a substantial expansion of the state and action spaces, which increases the per-step computational complexity and consequently makes each iteration of the fine-tuning more time-consuming. By comparison, the SRL requires a longer total training time compared to the PPO baseline as the pre-training and offline fine-tuning of SRL introduce a computational overhead. However, this increase is modest and the relative time increase shrinks as the problem scale grows. For the smallest-scale case of 5 DERs, the training time increases by 14\%. This percentage gradually decreases to 5\% for the large-scale case of 120 DERs. This trend can be attributed to the fact that the RL fine-tuning dominates the overall training duration as the state and action spaces expand.

The cost reduction of the SRL over the standard PPO on the test set is presented in Fig. \ref{fig_scalability}(b). The cost reduction is calculated as the percentage reduction in the average cumulative cost achieved by SRL over 10 independent trials relative to the PPO baseline. The result demonstrates that SRL achieves a substantial performance improvement across all scales. The cost reduction is particularly pronounced in the case of 30 DERs, where a 10\% increase in training time yields a cost reduction of over 10\%. However, a decline occurs in the cost reduction when the number of DERs exceeds 30. This phenomenon is likely a consequence of the curse of dimensionality. Specifically, the expansion of the state and action spaces in the current single-agent set-up leads to increasing sparsity of training data, which significantly reduces sample efficiency and makes it more difficult for the agent to improve its policy. Though the proposed SRL equips the agent with a reasonable base policy via pre-training, the curse of dimensionality restricts the exploration efficiency for both methods, thus narrowing the relative performance gap in larger-scale problems. Future work can investigate multi-agent reinforcement learning to alleviate this issue by factorizing the joint state-action space into decentralized components for individual agents, while maintaining global coordination through shared training signals or communication mechanisms.


\section{Conclusions}
\label{sec:conclusion}
This paper proposes a Supervised Reinforcement Learning (SRL) framework for DER control problems. This framework first establishes a baseline policy via supervised learning on demonstrations and subsequently fine-tunes it using reinforcement learning (RL). Furthermore, we propose a two-step fine-tuning process, comprising: an offline step in a simulated environment to refine the policy, and an online step in the actual environment to adapt the policy to real-world dynamics. Through comprehensive experiments, we demonstrate that the implementations based on the proposed framework achieve superior performance against all benchmarks. Notably, the proposed framework exhibits exceptional robustness, achieving high cost efficiency even when trained with low-quality demonstration data.




%

\bibliographystyle{IEEEtran}
\bibliography{ref_manuscript}

\end{document}